\documentclass[10pt]{article}

\usepackage{jmlr2e}
\usepackage[utf8]{inputenc}
\usepackage[T1]{fontenc}
\usepackage[english]{babel}
\usepackage{natbib}
\usepackage{listings}
\usepackage{xspace}
\usepackage{paralist}
\usepackage[nolist]{acronym}
\usepackage{tabularx}
\usepackage{csquotes}

\acrodef{ASR}{automatic speech recognition}
\acrodef{CI}{Continuous integration}
\acrodef{fMRI}{functional magnetic resonance imaging}
\acrodef{PCA}{principal component analysis}
\acrodef{PSD}{positive semidefinite}
\acrodef{SPD}{Symmetric positive definite}

\def\Title{Geomstats: A Python Package for Riemannian Geometry in Machine
Learning}

\newcommand*{\Toolbox}[1]{\textsf{#1}\xspace}
\newcommand*{\codeobj}[1]{\texttt{#1}\xspace}

\newcommand*{\Geomstats}{\textsc{Geomstats}\xspace}
\newcommand*{\Pymanopt}{\Toolbox{Pymanopt}}
\newcommand*{\Geoopt}{\Toolbox{Geoopt}}
\newcommand*{\McTorch}{\Toolbox{McTorch}}
\newcommand*{\PyRiemann}{\Toolbox{PyRiemann}}
\newcommand*{\PyQuaternion}{\Toolbox{PyQuaternion}}
\newcommand*{\PyGeometry}{\Toolbox{PyGeometry}}
\newcommand*{\TheanoGeometry}{\Toolbox{TheanoGeometry}}

\newcommand*{\GL}[1]{\mathrm{GL}(#1)}
\newcommand*{\SO}[1]{\mathrm{SO}(#1)}
\newcommand*{\SE}[1]{\mathrm{SE}(#1)}
\newcommand*{\SPD}[1]{\mathrm{SPD}(#1)}
\newcommand*{\sphere}[1]{\mathbb{S}^{#1}}
\newcommand*{\hyperbolic}[1]{\mathbb{H}^{#1}}

\newcommand*{\Href}[2]{\href{#1}{\texttt{#2}}}

\ShortHeadings{\Title}{The \Geomstats Team}
\firstpageno{1}

\begin{document}

\title{\Title}

\author{%
  \name Nina Miolane \email nmiolane@stanford.edu \\
  \name Alice Le Brigant \email alice.le-brigant@univ-paris1.fr \\
  \name Johan Mathe \email johan@froglabs.ai \\
  \name Benjamin Hou \email benjamin.hou11@imperial.ac.uk \\
  \name Nicolas Guigui \email nicolas.guigui@inria.fr \\
  \name Yann Thanwerdas \email yann.thanwerdas@inria.fr \\
  \name Stefan Heyder \email stefan.heyder@tu-ilmenau.de \\
  \name Olivier Peltre \email opeltre@gmail.com \\
  \name Niklas Koep \email niklas.koep@gmail.com \\
  \name Hadi Zaatiti \email hadi.zaatiti@irt-systemx.fr \\
  \name Hatem Hajri \email hatem.hajri@irt-systemx.fr \\
  \name Yann Cabanes \email yann.cabanes@gmail.com \\
  \name Thomas Gerald \email thomas.gerald@lip6.fr \\
  \name Paul Chauchat \email pchauchat@gmail.com \\
  \name Christian Shewmake \email cshewmake2@gmail.com \\
  \name Bernhard Kainz \email b.kainz@imperial.ac.uk \\
  \name Claire Donnat \email cdonnat@stanford.edu \\
  \name Susan Holmes \email susan@stat.stanford.edu \\
  \name Xavier Pennec \email xavier.pennec@inria.fr%
}

\editor{Francis Bach, David Blei and Bernhard Schölkopf}

\maketitle

\begin{abstract}
  We introduce \Geomstats, an open-source Python toolbox for computations and
  statistics on nonlinear manifolds, such as hyperbolic spaces, spaces of
  symmetric positive definite matrices, Lie groups of transformations, and many
  more.
  We provide object-oriented and extensively unit-tested implementations.
  Among others, manifolds come equipped with families of Riemannian metrics,
  with associated exponential and logarithmic maps, geodesics and parallel
  transport.
  Statistics and learning algorithms provide methods for estimation, clustering
  and dimension reduction on manifolds.
  All associated operations are vectorized for batch computation and provide
  support for different execution backends, namely NumPy, PyTorch and
  TensorFlow, enabling GPU acceleration.
  This paper presents the package, compares it with related libraries and
  provides relevant code examples.
  We show that \Geomstats provides reliable building blocks
  to foster research in differential geometry and statistics, and to
  democratize the use of Riemannian geometry in machine learning applications.
  The source code is freely available under the MIT license at
  \Href{http://geomstats.ai}{geomstats.ai}.
\end{abstract}

\section{Introduction}

Data on manifolds naturally arise in different fields. \textbf{Hyperspheres}
model directional data in molecular and protein biology \citep{Kent2005UsingStructure}, and are used through Principal Nested Spheres to model some aspects of 3D shapes \citep{Jung2012AnalysisSpheres, Hong2016}. Computations on
\textbf{hyperbolic spaces} arise for impedance density estimation
\citep{Huckemann2010MobiusEstimation}, geometric network comparison
\citep{Asta2014GeometricComparison} or the analysis of reflection coefficients
extracted from a radar signal \citep{Chevallier2015ProbabilityProcessing}.
\textbf{\ac{SPD} matrices} characterize data from
diffusion tensor imaging (DTI) \citep{Yuan2012} and \ac{fMRI}
\citep{Sporns2005TheBrain}.
Covariance matrices,
which are also \ac{SPD} matrices, appear in many fields like \ac{ASR} systems \citep{Shinohara2010}, image and video descriptors
\citep{Harandi2014}, or air traffic complexity representation
\citep{LeBrigant2018}.
The \textbf{Lie groups of transformations} $\SO{3}$ and
$\SE{3}$ appear naturally when dealing with articulated objects like the human
spine \citep{Arsigny:PHD:2006, Boisvert2006PrincipalModels}, or the pose of a
camera \citep{Kendall2017GeometricLearning, Hou2018ComputingGeometry}.
\textbf{Stiefel manifolds} are used to process video action data or to analyze
two vector-cardiograms in \citep{Chakraborty2019StatisticsApplications}.
\textbf{Grassmannians} appear in computer vision to perform video-based face
recognition and shape recognition \citep{Chellappa2008StatisticalPark}.
Statistics on \textbf{landmark spaces} are used in anthropology and many applied fields to describe
biological shapes \citep{Richtsmeier1992AdvancesMorphometrics}. More generally, Kendall shape spaces gave rise to a very important literature in shape statistics \citep{Dryden1998}. A variety of
applications use infinite-dimensional Riemannian geometry tools to study the
shapes of \textbf{discretized curves}, sampled e.g. from closed two or
three-dimensional curves defining the contours of organs in computational
anatomy \citep{Younes2012Spaces}.
We also find open curves embedded in any of
the previously mentioned manifolds that describe the temporal evolution of
physical phenomena, for example in a Lie group \citep{Celledoni2015} or in the
hyperbolic plane \citep{LeBrigant2017}.

Yet, the adoption of differential geometry computations has been inhibited by
the lack of a reference implementation.
Code sequences are often
custom-tailored for specific problems and are not easily reused.
Some Python packages do exist, but focus on optimization: \Pymanopt
\citep{Townsend2016Pymanopt:Differentiation}, \Geoopt
\citep{Becigneul2018RiemannianMethods, Kochurov2019Geoopt:Optim}, and
\McTorch \citep{Meghwanshi2018McTorchLearning}.
Others are dedicated
to a single manifold: \PyRiemann on \ac{SPD} matrices
\citep{Barachant2015PyRiemann:Interface}, \PyQuaternion on 3D
rotations \citep{Wynn2014PyQuaternions:Quaternions}, \PyGeometry on
spheres, toruses, 3D rotations and translations
\citep{Censi2012PyGeometry:Manifolds.}.
Lastly, others lack unit-tests and
continuous integration: \TheanoGeometry
\citep{Kuhnel2017ComputationalTheano}.
There is a need for an open-source
low-level implementation of differential geometry, and associated learning
algorithms for manifold-valued data.

We present \Geomstats, an open-source Python package of computations
and statistics on nonlinear manifolds. \Geomstats has three main
objectives:
\begin{inparaenum}[(i)]
  \item foster research in differential geometry and geometric statistics by
    providing low-level code to get intuition or test a theorem and a platform
    to share algorithms;
  \item democratize the use of geometric statistics by implementing
    user-friendly geometric learning algorithms using Scikit-Learn API; and
  \item provide educational support to learn \enquote{hands-on} differential
    geometry and geometric statistics, through its examples and visualizations.
\end{inparaenum}

\section{Implementation overview}\label{sec:geomstats}

The package \codeobj{geomstats} is organized into two main modules:
\codeobj{geometry} and \codeobj{learning}. \textbf{The module \codeobj{geometry}
implements concepts in Riemannian geometry} with an object-oriented approach.
Manifolds mentioned in the introduction are available as classes that inherit
from the base class \codeobj{Manifold}.
The base class \codeobj{RiemannianMetric} provides
methods such as the geodesic distance between two points, the exponential and
logarithm maps at a base point, etc.
As examples, \codeobj{HyperbolicMetric,
StiefelCanonicalMetric}, the Lie groups' \codeobj{InvariantMetrics}, or the
curves' \codeobj{L2Metric} and \codeobj{SRVMetric} \citep{Srivastava2011}
inherit from \codeobj{RiemannianMetric}.
Going beyond Riemannian geometry, the
class \codeobj{Connection} implements affine connections.
The implementation
uses automatic differentiation with autograd to allow computations on manifolds
when closed-form formulae do not exist.
The API of the module \codeobj{geometry}
uses terms from differential geometry, to foster contributions of researchers
from this field. \textbf{The module \codeobj{learning} implements statistics and
learning algorithms} for data on manifolds.
The code is object-oriented and
classes inherit from Scikit-Learn base classes and mixin:
\codeobj{BaseEstimator}, \codeobj{ClassifierMixin}, \codeobj{RegressorMixin}, etc.
This module provides classes such as \codeobj{FrechetMean}, \codeobj{KMeans},
\codeobj{TangentPCA}, for implementations of Fr\'echet mean estimators
\citep{Frechet1948}, $K$-means and \ac{PCA} designed for
manifold data.
The API of the module \codeobj{learning} follows Scikit-Learn's
API, being therefore user-friendly to machine learning researchers and
engineers.

\begin{figure}[!h]
  \centering
  \includegraphics[width=122pt]{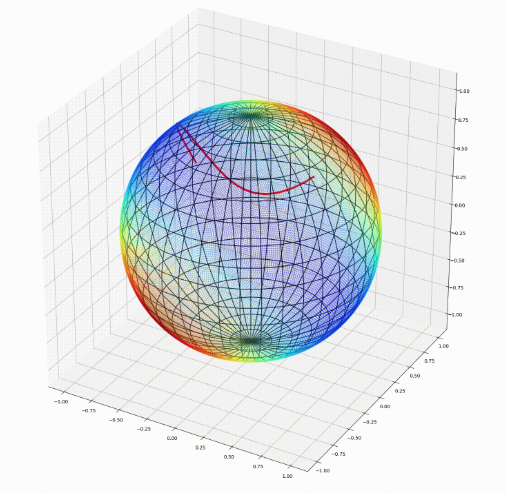}
  \includegraphics[width=150pt]{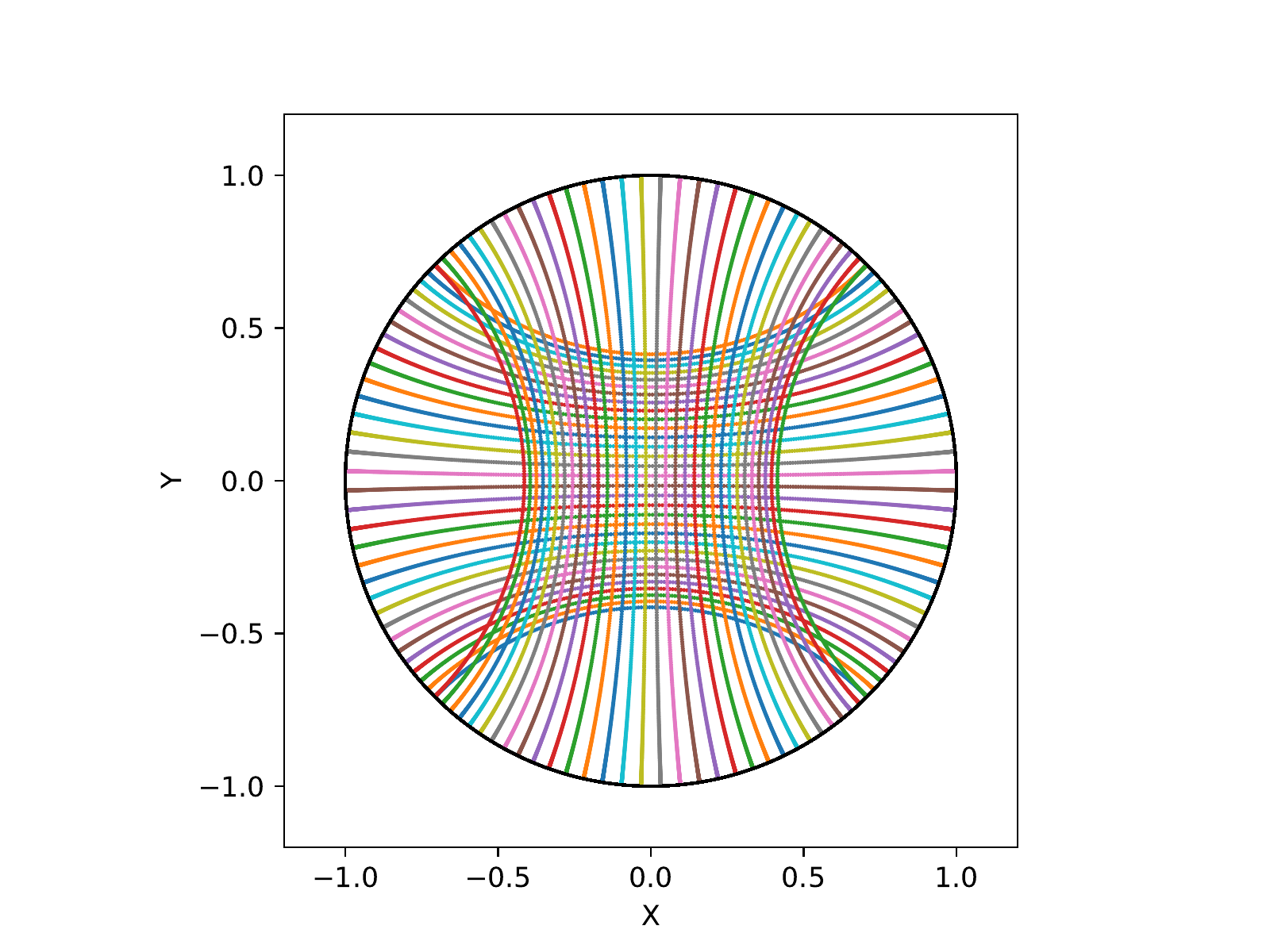}
  \includegraphics[width=120pt]{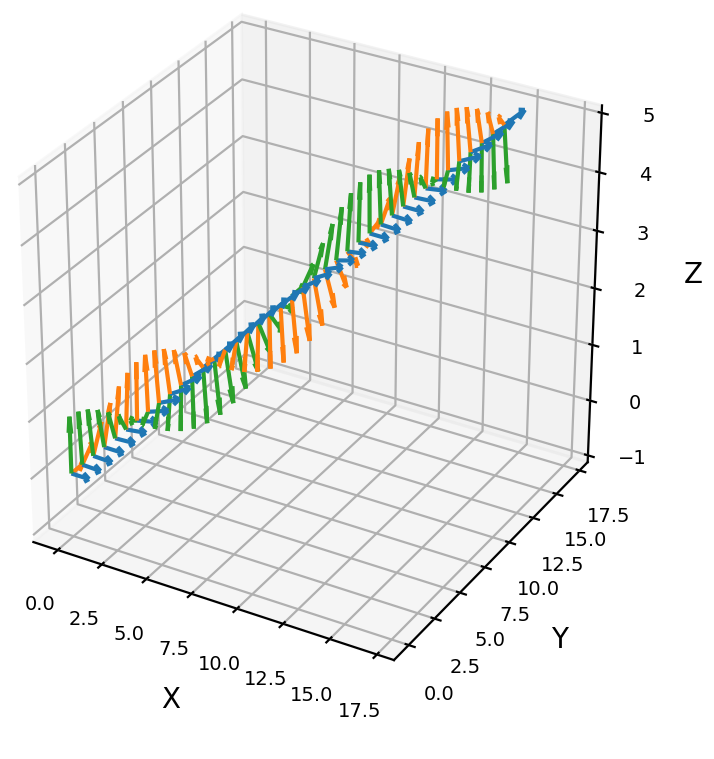}
  \caption{%
    Left: Minimization of a scalar field on the sphere $\sphere{2}$ using
    Riemannian gradient descent. \newline
    Middle: Regular geodesic grid on the hyperbolic space $\hyperbolic{2}$ in
    Poincar\'e disk representation. \newline
    Right: Geodesic on the Lie group $\SE{3}$ for the canonical left-invariant
    metric. \newline
    These and more examples are available at
    \Href{http://geomstats.ai}{geomstats.ai}.%
  }
  \label{fig:visualization}
\end{figure}

The code follows international standards for readability and ease of
collaboration, is vectorized for batch computations, undergoes unit-testing
with continuous integration using Travis, relies on TensorFlow and PyTorch
backends for GPU acceleration, and is partially ported to R.
The package comes
with a \codeobj{visualization} module to provide intuitions on differential
geometry, see Figure~\ref{fig:visualization}.
The GitHub repository at
\Href{https://github.com/geomstats/geomstats}{github.com/geomstats/geomstats}
offers a convenient way to ask for help or request features by raising issues.
The website \Href{https://geomstats.github.io}{geomstats.github.io}
provides documentation for users and important guidelines for those wishing to
contribute to the project.

\section{Comparison and interaction with existing packages}

This section compares the package \codeobj{geomstats} with related Python
implementations on differential geometry and learning.
Table~\ref{tab:geometry}
compares the geometric operations and Table~\ref{tab:eng} compares the
engineering infrastructures.

The library \TheanoGeometry \citep{Kuhnel2017ComputationalTheano} is
the most closely related to \Geomstats and provides nonlinear statistics and
stochastic equations on Riemannian manifolds.
The differential geometric
tensors are computed with automatic differentiation.
However, this library
does not provide statistical learning algorithms and lacks engineering
maintenance.
Then, there are several packages that focus on optimization on
Riemannian manifolds. \Pymanopt
\citep{Townsend2016Pymanopt:Differentiation} computes gradients and
Hessian-vector products on Riemannian manifolds with automatic differentiation,
and provides the following solvers: steepest descent, conjugate gradients,
the Nelder-Mead algorithm, particle swarm optimization and the Riemannian trust
regions.
\Geoopt \citep{Kochurov2019Geoopt:Optim} focuses on stochastic
adaptive optimization on Riemannian manifolds, for machine learning problems.
The library provides stochastic solvers, stochastic gradient descent and
Adam, as well as the following samplers: Stochastic Gradient Langevin Dynamics,
Hamiltonian Monte-Carlo, Stochastic Gradient Hamiltonian Monte-Carlo.
Lastly, \McTorch \citep{Meghwanshi2018McTorchLearning} provides optimization on
Riemannian manifold for deep learning by adding a \enquote{Manifold} parameter
to PyTorch's network layers and optimizers.
The library provides the following
solvers: stochastic gradient descent, AdaGrad and conjugate gradients.
As these libraries focus on optimization, they substitute potentially
computationally expensive operations by practical proxies, for example, by
replacing exponential maps by so-called retractions.
However, they are less
modular than \Geomstats in terms of the Riemannian geometry.
For example, each
manifold comes with a single Riemannian metric, in contrast to \Geomstats where
families of Riemannian metrics are implemented.
Furthermore, they do not
provide statistical learning algorithms.

The optimization libraries are complementary to \Geomstats and interact
easily with it. \Geomstats provides low-level implementations of
Riemannian geometry that can be used to define optimization costs.
In turn, an
optimization library can provide an efficient solver to use within the
implementation of \Geomstats' learning algorithms.
An example of such interactions, between \Pymanopt and \Geomstats can be found
in \Geomstats' \texttt{examples} folder.

\begin{table}
  \footnotesize
  \centering
  \begin{tabularx}{\linewidth}{|l|X|X|}
    \hline
    & \textbf{Manifolds} & \textbf{Geometry} \\ \hline
    \textbf{Pymanopt}
    & Euclidean manifold, symmetric matrices,
    sphere, complex circle, $\mathrm{SO}(n)$, Stiefel,
    Grassmannian, oblique manifold, $\SPD{n}$,
    elliptope, fixed-rank \acs{PSD} matrices
    & Exponential and logarithmic maps, retraction, vector transport,
    \codeobj{egrad2rgrad}, \codeobj{ehess2rhess}, inner product, distance, norm
    \\ \hline

    \textbf{Geoopt}
    & Euclidean manifold, sphere, Stiefel, Poincar\'e ball
    & Same as \Pymanopt \\ \hline

    \textbf{McTorch}
    & Stiefel, $\SPD{n}$
    & Same as \Pymanopt \\ \hline

    \textbf{TheanoGeometry}
    & Sphere, ellipsoid, $\SPD{n}$, Landmarks, $\GL{n}$, $\SO{n}$, $\SE{n}$
    & Inner product, exponential and logarithmic maps, parallel transport,
    Christoffel symbols, Riemann, Ricci and scalar curvature, geodesics,
    Fr\'echet mean \\ \hline

    \textbf{Geomstats}
    & Euclidean manifold, Minkowski and hyperbolic space, sphere, $\SO{n}$,
    $\SE{n}$, $\GL{n}$, Stiefel, Grassmannian, $\SPD{n}$, discretized curves,
    Landmarks
    & Levi-Civita connection, parallel transport, exponential and logarithmic
    maps, inner product, distance, norm, geodesics, invariant metrics \\ \hline
  \end{tabularx}
  \caption{Comparison of libraries in terms of geometric operations}
  \label{tab:geometry}
\end{table}

\begin{table}
  \footnotesize
  \centering
  \begin{tabularx}{\linewidth}{|l|l|X|}
    \hline
    & \textbf{Backends}
    & \textbf{\acf{CI} and coverage}                                                       \\ \hline

    \textbf{Pymanopt}
    & Autograd, PyTorch, TensorFlow, Theano
    & CI, coverage 85\% \\ \hline

    \textbf{Geoopt}
    & PyTorch
    & Not disclosed \\ \hline

    \textbf{McTorch}
    & PyTorch
    & CI, coverage 84\% \\ \hline

    \textbf{TheanoGeometry}
    & Theano
    & No CI, no unit tests \\ \hline

    \textbf{Geomstats}
    & NumPy, PyTorch, TensorFlow
    & CI, coverage 89\% (NumPy), 45\% (TensorFlow), 47\% (PyTorch) \\ \hline
  \end{tabularx}
  \caption{Comparison of code infrastructure}
  \label{tab:eng}
\end{table}

\section{Usage: Examples of Learning on Riemannian manifolds}

The folder \texttt{examples} provides sample codes to get started with
\Geomstats.
More involved \Geomstats applications can be found at:
\Href{https://github.com/geomstats/applications}{github.com/geomstats/applications}.
This section illustrates the use of \Geomstats to generalize learning
algorithms to Riemannian manifolds.
The following code snippet illustrates the use of $K$-means on the hypersphere.

\begin{lstlisting}[language=python]
    sphere = Hypersphere(dimension=5)
    clustering = OnlineKMeans(metric=sphere.metric, n_clusters=4)
    clustering = clustering.fit(data)
\end{lstlisting}

\noindent The following code snippet shows the use of tangent \ac{PCA} on the
3D rotations.

\begin{lstlisting}[language=python]
    so3 = SpecialOrthogonal(n=3)
    metric = so3.bi_invariant_metric
    tpca = TangentPCA(metric=metric, n_components=2)
    tpca = tpca.fit(data, base_point=metric.mean(data))
    tangent_projected_data = tpca.transform(data)
\end{lstlisting}

All geometric computations are performed behind the scenes.
The user only needs
a high-level understanding of Riemannian geometry.
Each algorithm can be used
with any of the manifolds and metric implemented in the package.

\section*{Acknowledgments}

This work is partially supported by the National Science Foundation, grant NSF
DMS RTG 1501767, the Inria-Stanford associated team GeomStats, and the European
Research Council (ERC) under the European Union's Horizon 2020 research and
innovation program (grant agreement G-Statistics No. 786854).

\setlength{\bibsep}{1.5pt plus 0.1ex}
\footnotesize
\bibliography{main}

\end{document}